%% file: main.tex
\RequirePackage[svgnames,table]{xcolor}

\documentclass{article}
\usepackage{iclr2026_conference,times}
\usepackage{xpeng}
\setxpenglab{XPENG Robotics}
\setxpengreport{XPENG Technical Report}
\providecommand{\runningtitle}[1]{}
\providecommand{\keywords}[1]{}

\usepackage{xcolor}

\usepackage[normalem]{ulem} %
\usepackage{color}          %

\input{preamble}   %
\input{command}    %

\usepackage{fontawesome5}
\usepackage{graphicx}
\usepackage{hyperref}   %
\usepackage{placeins}   %
\usepackage[normalem]{ulem}
\usepackage{xcolor}
\title{IronMind: Scaling Humanoid Dexterous Manipulation \\ via Camera-Space Ego-Centric Pretraining}

\runningtitle{IronMind: Scaling Humanoid Dexterous Manipulation via Camera-Space Ego-Centric Pretraining}

\keywords{vision-language-action models, humanoid manipulation, egocentric data, mixture of transformers, scaling}

\author{%
World Model Team, XPENG Robotics\\[0.35em]
{\normalfont\small\faGithub\ Project: \href{https://xpeng-robotics.github.io/ironmind/}{\nolinkurl{https://xpeng-robotics.github.io/ironmind/}}}
}

\hypersetup{colorlinks=true, linkcolor=blue!50!black, citecolor=blue!50!black,
            urlcolor=blue!50!black}

\begin{document}

\maketitle

\begin{abstract}
\input{sections/01_abstract/abstract}
\end{abstract}

\input{sections/02_introduction/introduction}
\input{sections/03_related_work/related_work}
\input{sections/04_method/overview}
\input{sections/04_method/data}
\input{sections/04_method/model}
\input{sections/04_method/training}

\input{sections/05_experiments/experiments}

\input{sections/06_conclusion/limitations}

\input{sections/06_conclusion/conclusion}

\bibliographystyle{plainnat}
\bibliography{main}

\section*{Contributors}
\label{sec:contributors}

Huimin Pan\textsuperscript{1*},
Yufan Ren\textsuperscript{1*\dag},
Kunpeng Song\textsuperscript{1*},
Siyang Wang\textsuperscript{1*},
Xiwen Zhang\textsuperscript{1*},
Xiaoyun Hu\textsuperscript{1,2\S},
Zhuoxu Duan\textsuperscript{1,3\S},
Hanrui Zheng\textsuperscript{1,4\S},
Jialeng Ni\textsuperscript{1,5\S},
Nathan Zhao\textsuperscript{1},
Sibo Ma\textsuperscript{1,6\S},
Zhenxuan Fan\textsuperscript{1,7\S},
Zhongyang Che\textsuperscript{1},
Danny Bao\textsuperscript{1},
Jiacheng Wei\textsuperscript{1},
Jerry Bai\textsuperscript{1},
Xiaoyu Yue\textsuperscript{1},
Xiaoyang Guo\textsuperscript{1},
Chenyi Chen\textsuperscript{1\ddag}

\vspace{1.5mm}
\noindent
\textsuperscript{*}\,Equal contribution, listed alphabetically by last name.\quad
\textsuperscript{\dag}\,Project lead.\quad
\textsuperscript{\ddag}\,Supervision.\quad
\textsuperscript{\S}\,Work done during an internship at XPENG Robotics.

\vspace{1mm}
\noindent
\textsuperscript{1}\,XPENG Robotics\quad
\textsuperscript{2}\,Tongji University\quad
\textsuperscript{3}\,Rensselaer Polytechnic Institute\quad
\textsuperscript{4}\,University of Science and Technology of China\quad
\textsuperscript{5}\,University of Michigan\quad
\textsuperscript{6}\,University of British Columbia\quad
\textsuperscript{7}\,Zhejiang University

\section*{Acknowledgements}

We thank members of the XPENG Robotics team---Zike Cheng, Ruixuan Zhao, Mengen Luo, Luoyu Bai, Ying Zhang, Haoyue Liu, Peipeng Chen, Ruyi Li, Yizhe Liu, Yongqi Meng, Yinggan Xu, Xi Chen, Bin Shen, and Yuying Ge---for insightful discussions and deployment support, and Zhenglong Du, Hongmei Gan, and Weiqi Liang for assistance with teleoperation and data collection.

\end{document}

%% file: preamble.tex
\usepackage{graphicx}
\usepackage{float,epstopdf}
\usepackage{bbm}

\usepackage{microtype}

\usepackage{natbib}
\setcitestyle{square}

\usepackage{subcaption}
\usepackage{booktabs}

\usepackage{amsmath}
\usepackage{amssymb}
\usepackage{mathtools}
\usepackage{amsthm}
\usepackage{dsfont}
\usepackage{multicol}
\usepackage{makecell}
\usepackage{multirow} 
\usepackage{amsfonts} 
\usepackage{mathrsfs}
\usepackage{enumitem}
\usepackage{pgfplotstable}
\pgfplotsset{compat=1.18}
\usepackage{lipsum}		%

\usepackage{microtype}
\usepackage{graphicx}
\usepackage{booktabs} %
\usepackage[table]{xcolor}
\usepackage{arydshln}
\usepackage[normalem]{ulem} %

\usepackage{cases}
\usepackage{wrapfig}

\usepackage{url}

\usepackage{thmtools}
\usepackage{thm-restate}
\usepackage{tabu}

\definecolor{huskypurple}{HTML}{4B2E83}

\usepackage{titletoc}

\usepackage{listings}
\lstdefinestyle{promptstyle}{
  basicstyle=\ttfamily\footnotesize,
  breaklines=true,
  breakautoindent=false,
  breakindent=0pt,
  postbreak=\mbox{\textcolor{gray}{$\hookrightarrow$}\space},
  columns=fullflexible,
  keepspaces=true,
  frame=single,
  framesep=5pt,
  xleftmargin=6pt,
  xrightmargin=6pt,
  aboveskip=8pt,
  belowskip=8pt,
  showstringspaces=false,
}

\newif\ifreview
\reviewfalse

\ifreview
  \usepackage[mathlines]{lineno}
  \newcommand*\patchAmsMathEnvironmentForLineno[1]{%
    \expandafter\let\csname old#1\expandafter\endcsname\csname #1\endcsname
    \expandafter\let\csname oldend#1\expandafter\endcsname\csname end#1\endcsname
    \renewenvironment{#1}%
      {\linenomath\csname old#1\endcsname}%
      {\csname oldend#1\endcsname\endlinenomath}%
  }
  \newcommand*\patchBothAmsMathEnvironmentsForLineno[1]{%
    \patchAmsMathEnvironmentForLineno{#1}%
    \patchAmsMathEnvironmentForLineno{#1*}%
  }
  \AtBeginDocument{%
    \patchBothAmsMathEnvironmentsForLineno{equation}%
    \patchBothAmsMathEnvironmentsForLineno{align}%
    \patchBothAmsMathEnvironmentsForLineno{flalign}%
    \patchBothAmsMathEnvironmentsForLineno{alignat}%
    \patchBothAmsMathEnvironmentsForLineno{gather}%
    \patchBothAmsMathEnvironmentsForLineno{multline}%
    \linenumbers
  }
\fi

\usepackage{needspace}   %

%% file: command.tex
\makeatletter
\def\munderbar#1{\underline{\sbox\tw@{$#1$}\dp\tw@\z@\box\tw@}}
\makeatother

\AddToHook{cmd/appendix/before}{%
  \setcounter{axiom}{0}%
}

\newcommand{\be}{\begin{equation}}
\newcommand{\ee}{\end{equation}}
\newcommand{\bea}{\begin{equation*}\begin{aligned}}
\newcommand{\eea}{\end{aligned}\end{equation*}}

%% file: sections/01_abstract/abstract.tex
Egocentric human video offers a scalable data source for dexterous manipulation, yet using it to train humanoid robots presents two challenges: 
(1) an \emph{embodiment gap}, as human hands differ structurally from robot end-effectors and low-cost egocentric recordings lack the torso kinematics required by conventional retargeting; and 
(2) \emph{heterogeneous data quality}, including noisy hand-pose tracking and weakly aligned text annotations.
We introduce \textbf{IronMind}, a vision-language-action (VLA) model that uses egocentric human video and heterogeneous robot data to pretrain policies for humanoid dexterous manipulation. 
To bridge the embodiment gap, IronMind bypasses explicit body-retargeting by using a \emph{camera-space action representation}, the native reference space of egocentric video, and semantically aligning robot and human action dimensions. 
To mitigate data noise, we curate a pretraining corpus totaling more than 10,000 hours of egocentric human video and heterogeneous robot data using rule-based filtering, atomic task re-annotation, and per-frame quality weighting.
Architecturally, IronMind uses a Mixture-of-Transformers policy that couples vision-language understanding with a flow-matching action expert. 
We also explore optional world-prior supervision from semantic features, spatial-geometric representations, and temporal world-model predictions.
Across total pretraining budgets from 250 to 10,000 hours, validation loss decreases approximately log-linearly with data scale.
Larger pretraining budgets also improve out-of-distribution real-robot manipulation after post-training: across six challenging tasks with unseen objects, affordances, and reasoning prompts, the 10,000-hour model achieves a $55.0\%$ success rate, compared with at most 11.7\% for every pretraining budget up to 5,000 hours and 5.0\% without pretraining. 
At the same pretraining budget, the camera-space action representation also outperforms the torso-frame baseline.
Together, these findings support pretraining with a camera-space action representation on large-scale human egocentric data as a scalable foundation for humanoid robot manipulation.

%% file: sections/02_introduction/introduction.tex
\section{Introduction}
\label{sec:introduction}

\begin{figure}[t]
  \centering
  \includegraphics[width=\linewidth]{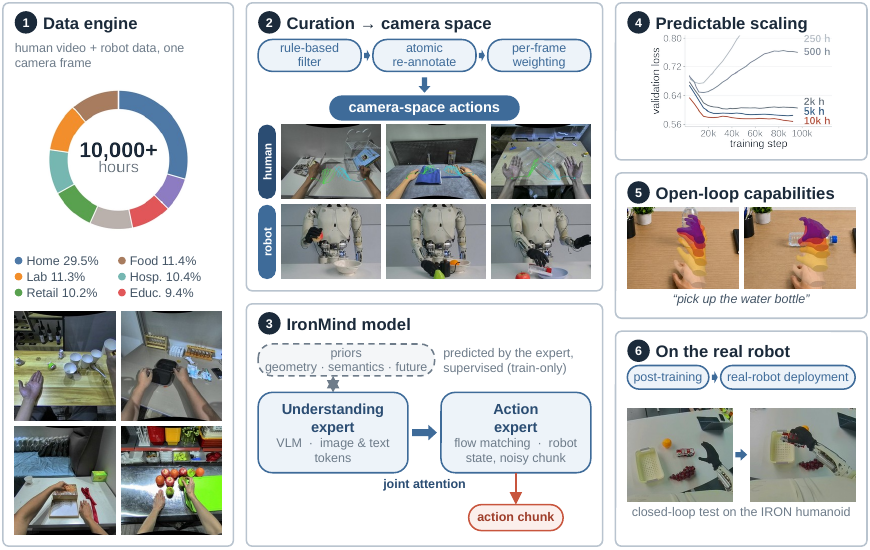}
  \caption{\textbf{Overview of IronMind.} 
  (1) Pretraining Corpus: Over 10,000 hours in total, combining egocentric human video with heterogeneous robot data from non-target embodiments.
  (2) Curation \& Representation: Data filtered, re-annotated, and unified into a shared camera-space action representation.
  (3) Architecture: A Mixture-of-Transformers policy with flow-matching action experts and optional multimodal priors.
  (4) Scaling: Validation loss scales log-linearly with pretraining hours without saturation up to 10,000 hours.
  (5) Zero-shot Evaluation: Open-loop trajectory prediction from a single unseen image and instruction.
  (6) Real-Robot Deployment: Closed-loop policy execution on IRON-R01.}
  \label{fig:teaser}
\end{figure}

Robot manipulation has advanced rapidly~\citep{chi2023diffusion,octo2024,liu2024rdt}, driven by
vision-language-action (VLA) models built on pretrained vision-language
backbones that follow open-vocabulary instructions and generalize
semantically across objects and scenes~\citep{brohan2023rt2,kim2024openvla,black2024pi0,nvidia2025gr00t},
achieving high success rates in structured tabletop and household tasks~\citep{intelligence2025pi05,jiang2025galaxea}. However, scaling these models to general dexterous manipulation—defined as a unified policy that handles contact-rich, high-degree-of-freedom hand--object interactions—remains limited by data availability. Demonstrations of dexterous manipulation on real robots are slow and expensive to collect~\citep{oneill2024openx,agibotworld2025}, and robotic manipulation lacks a web-scale data corpus.

Egocentric human video provides a low-cost data source for manipulation pretraining~\citep{li2025vitra,egoscale2026,egowam2026}, with 3D hand poses and camera motion estimated from head-mounted recordings. However, using egocentric video presents two primary challenges:
(1) an \emph{embodiment gap}: human hands differ structurally from robot end-effectors in kinematics and joint count, making direct retargeting inaccurate; and
(2) a \emph{quality gap}: hand-motion estimates contain tracking noise, text prompts are weakly aligned with the corresponding actions, and uncurated video includes segments without manipulation, reducing data efficiency~\citep{lin2024datascaling,belkhale2023dataquality,hejna2024remix}.
Addressing these two gaps is critical for effectively pretraining humanoid robot policies.

\begingroup\hbadness=10000
In this paper, we introduce \textbf{IronMind}, a vision-language-action model pretrained with a camera-space action representation on large-scale egocentric human video for humanoid dexterous manipulation (Figure~\ref{fig:teaser}). 
To address the embodiment gap, IronMind avoids human-to-robot body retargeting. Egocentric video supports the estimation of camera-space trajectories (camera motion via SLAM and hand motion relative to camera space) rather than the wearer's global body pose. Projecting these trajectories into a robot's joint space requires assumptions about unobserved body kinematics (such as a rigid spine model), introducing label noise. IronMind instead expresses human and calibrated robot trajectories in the same camera-space action representation and aligns action dimensions semantically across embodiments~\citep{qwenrobotmanip2026}. 
To address the quality gap, a multi-stage curation pipeline applies rule-based filtering, atomic task re-annotation, and per-frame quality weighting to the pretraining video data.

Furthermore, dexterous manipulation requires predicting scene dynamics during hand--object contact. IronMind optionally incorporates world-model priors during pretraining by using learnable query tokens to condition a frozen video generation model~\citep{wan2025} for future-frame prediction~\citep{internvla_a15_2026}. This supervision is combined with semantic visual features~\citep{simeoni2025dinov3} and spatial-geometric representations~\citep{lin2025depthanything3}. 
The policy architecture uses a Mixture-of-Transformers~\citep{liang2024mot} with separate vision-language and flow-matching action experts~\citep{lipman2023flow,black2024pi0} linked via shared attention, preserving vision-language capabilities during action learning. All world-model prediction modules are removed after pretraining. This leaves only the learned query tokens and avoids auxiliary teacher-model computation at inference. Model performance is evaluated using camera-space open-loop metrics, policy-in-the-loop rollouts in held-out scenes, and real-robot testing.
\par\endgroup

We summarize our main contributions as follows:

\begin{enumerate}
  \item \textbf{A camera-space pretraining recipe for humanoid manipulation.} We use a camera-space action representation as a unified human--robot interface, combining semantic alignment of action dimensions across embodiments with a multi-stage curation pipeline (rule-based filtering, atomic re-annotation, per-frame quality weighting) to process a total of over 10,000 hours of egocentric human video and heterogeneous robot data (Section~\ref{sec:data}). Validation loss decreases log-linearly with the pretraining data budget.
  \item \textbf{A VLA architecture with optional multimodal world priors.} We introduce a dual-expert Mixture-of-Transformers backbone with flow-matching action decoding and explore optional world-prior supervision through future-frame prediction using a video generation model alongside semantic~\citep{simeoni2025dinov3} and spatial-geometric representations~\citep{lin2025depthanything3}, without running the auxiliary teacher models at inference (Sections~\ref{sec:model}--\ref{sec:training}).
  \item \textbf{An open-to-closed-loop evaluation suite.} We present an evaluation benchmark that combines camera-space open-loop approach and grasping metrics, policy-in-the-loop rollouts in unseen scenes, and real-robot evaluation protocols to assess prediction quality, sequential drift, and real-robot execution (Section~\ref{sec:experiments}).
  \item \textbf{Empirical validation on robots.} On a real humanoid robot, larger pretraining budgets yield substantial improvements across six challenging out-of-distribution tasks with unseen objects, affordances, and reasoning prompts. The 10,000-hour model achieves a $55.0\%$ success rate, representing a 43-percentage-point gain over smaller pretraining budgets (average success rates are at most $11.7\%$ for all budgets up to 5,000 hours); at the same budget, the camera-space action representation also outperforms the torso-frame baseline ($55.0\%$ vs. $26.7\%$) (Section~\ref{sec:experiments}).
\end{enumerate}

%% file: sections/03_related_work/related_work.tex
\section{Related Work}
\label{sec:related}

Building robot policies on pretrained vision-language backbones~\citep{bai2025qwen25vl,qwenteam2026qwen35} has become a standard approach to generalist manipulation. Early architectures cast robot control as the prediction of discrete text tokens~\citep{brohan2023rt2,kim2024openvla}, with later work improving efficiency through learned action tokenization~\citep{pertsch2025fast,galaxea2026g05}. Other designs use continuous action heads based on diffusion or flow matching~\citep{chi2023diffusion,octo2024,liu2024rdt,black2024pi0}. 
These approaches underpin a range of large-scale VLA systems for real-world robot control, including the $\pi$ series~\citep{intelligence2025pi05,intelligence2025pistar06,intelligence2026pi07}, GR00T~N1~\citep{nvidia2025gr00t}, Galaxea G0/G0.5~\citep{jiang2025galaxea,galaxea2026g05}, Qwen-VLA and Qwen-RobotManip~\citep{qwenvla2026,qwenrobotmanip2026}, LingBot-VLA~\citep{lingbotvla2026}, InternVLA-A1/A1.5~\citep{internvla_a1_2026,internvla_a15_2026}, Hy-Embodied-0.5~\citep{hyembodied2026}, and Capek~0.5~\citep{capek2026}, with related formulations also explored in autonomous driving~\citep{nvidia2025alpamayo,automot2026}. 
Parallel work investigates unified generation-action interfaces~\citep{wang2025univla,bagelvla2026}, adaptive reasoning~\citep{lin2025onetwovla,motvla2025}, memory mechanisms~\citep{torne2026mem,mapvla2025}, and minimal policy formulations~\citep{goyal2025vla0}. IronMind builds on these advances, focusing on scalable data representations and a multi-stage pretraining recipe for learning humanoid manipulation from egocentric human video.

\textbf{Learning manipulation from egocentric human data.} Large-scale egocentric
corpora~\citep{grauman2022ego4d,grauman2024egoexo4d,damen2022rescaling,goyal2017something} provide manipulation experience at a scale difficult to achieve with robot fleets alone, while recent datasets add accurate 3D hand-pose annotations~\citep{hoque2025egodex}. Building on these resources, systems such as VITRA, EgoScale, and EgoWAM demonstrate the effectiveness of egocentric pretraining at progressively larger scales~\citep{li2025vitra,egoscale2026,egowam2026}. 
A central design choice is how to represent actions for transfer from humans to robots. Existing approaches include latent actions learned from video~\citep{ye2024lapa,beingh07_2026}, explicit camera-space hand trajectories~\citep{li2025vitra}, relative wrist motion retargeted into robot joint space prior to training~\citep{egoscale2026}, robot trajectories synthesized via cross-domain translation~\citep{ego2robot2026,translation2026,simdex2026}, and joint human--robot representations~\citep{harpvla2026,lap2026,dypesvla2026}. 
A complementary approach avoids explicit conversion between embodiments by retaining data from heterogeneous sources in their native action spaces and aligning their action dimensions semantically~\citep{qwenrobotmanip2026,joyai_ra05_2026}. Beyond action representation, growing evidence highlights the importance of data quality and diversity alongside raw data volume~\citep{lin2024datascaling,belkhale2023dataquality,hejna2024remix}, while specialized tools facilitate data conversion~\citep{joyai_sim2026}. 
IronMind follows the same principle of avoiding body retargeting: human demonstrations use a camera-space action representation, while robot trajectories are transformed into the same camera space using directly measurable calibration parameters. Remaining differences in action dimensions are addressed through semantic alignment~\citep{qwenrobotmanip2026}, enabling a single output head to learn from both human and robot action supervision without body retargeting or a learned latent bottleneck.

Integrating video prediction with control is an active area of policy learning. Existing approaches include jointly modeling physical dynamics and actions~\citep{dit4dit2026,lingbotwa2026,worldsinonedemo2026,wei2026xpace}, using predicted future frames as policy context~\citep{dreamzero2026,fastwam2026,omega0_2026}, distilling latent world representations into the policy network~\citep{vlajepa2026,dial2026,chen2026unit}, scaling world-action pretraining on egocentric video corpora~\citep{beingh07_2026,egowam2026}, and building omnimodal world models~\citep{nvidia2026cosmos3}. 
Two findings from prior work guide our design. First, aligning representations of future states with action supervision helps make them useful for policy execution~\citep{agra2026,dial2026}. Second, pretrained video generation models encode rich priors about 3D dynamics and spatial structure~\citep{genknowspace2026,omniview2025,peebles2023dit,wan2025}. 
The closest architectural precedent is InternVLA-A1.5, which encodes task-relevant future visual states in latent foresight tokens supervised by a frozen pretrained video generator. This design incorporates dynamics priors without requiring pixel-level video synthesis at deployment~\citep{internvla_a15_2026}. IronMind adopts this approach, using world models solely to provide priors during training while leaving the video generator inactive at inference. We combine these predictive representations with semantic and geometric feature distillation~\citep{simeoni2025dinov3,lin2025depthanything3} within a Mixture-of-Transformers backbone~\citep{liang2024mot}, an architecture that combines expert-specific parameters with shared attention and has been applied to multimodal learning and embodied control~\citep{motvla2025,automot2026}.

\textbf{Scaling and evaluation for robot learning.} Scaling laws are well established for language models~\citep{kaplan2020scaling}; corresponding relationships in robot learning are being investigated through cross-embodiment datasets~\citep{oneill2024openx,agibotworld2025}, empirical studies of dataset size, composition, and quality~\citep{lin2024datascaling,hejna2024remix,belkhale2023dataquality,lin2026systematic}, and analyses of action chunking strategies~\citep{lazzati2026chunking,zhao2023aloha}. Complementary post-training methods, such as reinforcement learning and physical interventions, further enhance real-world policy performance~\citep{xu2026rltoken,wcm2026,intelligence2025pistar06,rove2026}.
Evaluation practices, however, remain fragmented: studies often rely on isolated offline metrics or success rates on a limited set of real-robot tasks, a limitation systematically analyzed by VIGIL~\citep{vigil2026}. To address this fragmentation, we introduce an open-to-closed-loop evaluation suite that combines camera-space open-loop approach and grasp metrics, policy-in-the-loop rollouts in held-out scenes, and standardized real-robot testing protocols. Organized around an explicit taxonomy of capabilities, the suite supports systematic reporting of both successes and failure modes.

%% file: sections/04_method/overview.tex
\section{IronMind: The Framework}
\label{sec:overview}

IronMind consists of three core components, as illustrated in Figure~\ref{fig:teaser}.
The \textbf{pretraining data engine} (Section~\ref{sec:data}) processes diverse egocentric human videos and heterogeneous robot demonstrations to address differences in embodiment and data quality. It uses a shared camera-space action representation for trajectories from different embodiments and applies a multi-stage curation pipeline to filter data affected by tracking failures, dropped frames, and kinematically invalid hand motions, while refining inaccurate text annotations.
The \textbf{model architecture} (Section~\ref{sec:model}) combines a vision-language-action (VLA) policy with optional multimodal priors, using a Mixture-of-Transformers~\citep{liang2024mot,internvla_a15_2026} that links a vision-language expert and a flow-matching action expert~\citep{lipman2023flow,black2024pi0} via shared attention. During training, the backbone can receive auxiliary supervision from semantic features~\citep{simeoni2025dinov3}, spatial-geometric representations~\citep{lin2025depthanything3}, and future-frame predictions from a frozen video generation model~\citep{wan2025}; the auxiliary prediction heads are removed at inference, leaving only a compact set of learned query tokens and avoiding the cost of running the auxiliary prediction branches during deployment.
The \textbf{training recipe} (Section~\ref{sec:training}) consists of large-scale pretraining with the camera-space action representation, followed by post-training on a small curated dataset of real-robot teleoperation demonstrations, with robot action trajectories transformed into the same camera space using extrinsic calibration.
Together, these components maintain a consistent action interface from heterogeneous pretraining data through adaptation and deployment on the target robot.

%% file: sections/04_method/data.tex
\subsection{Pretraining data engine}
\label{sec:data}

\begin{figure}[t]
  \centering
  \includegraphics[width=1.0\linewidth,
                    trim=2.2bp 136bp 2.65bp 55.9bp, clip]
                  {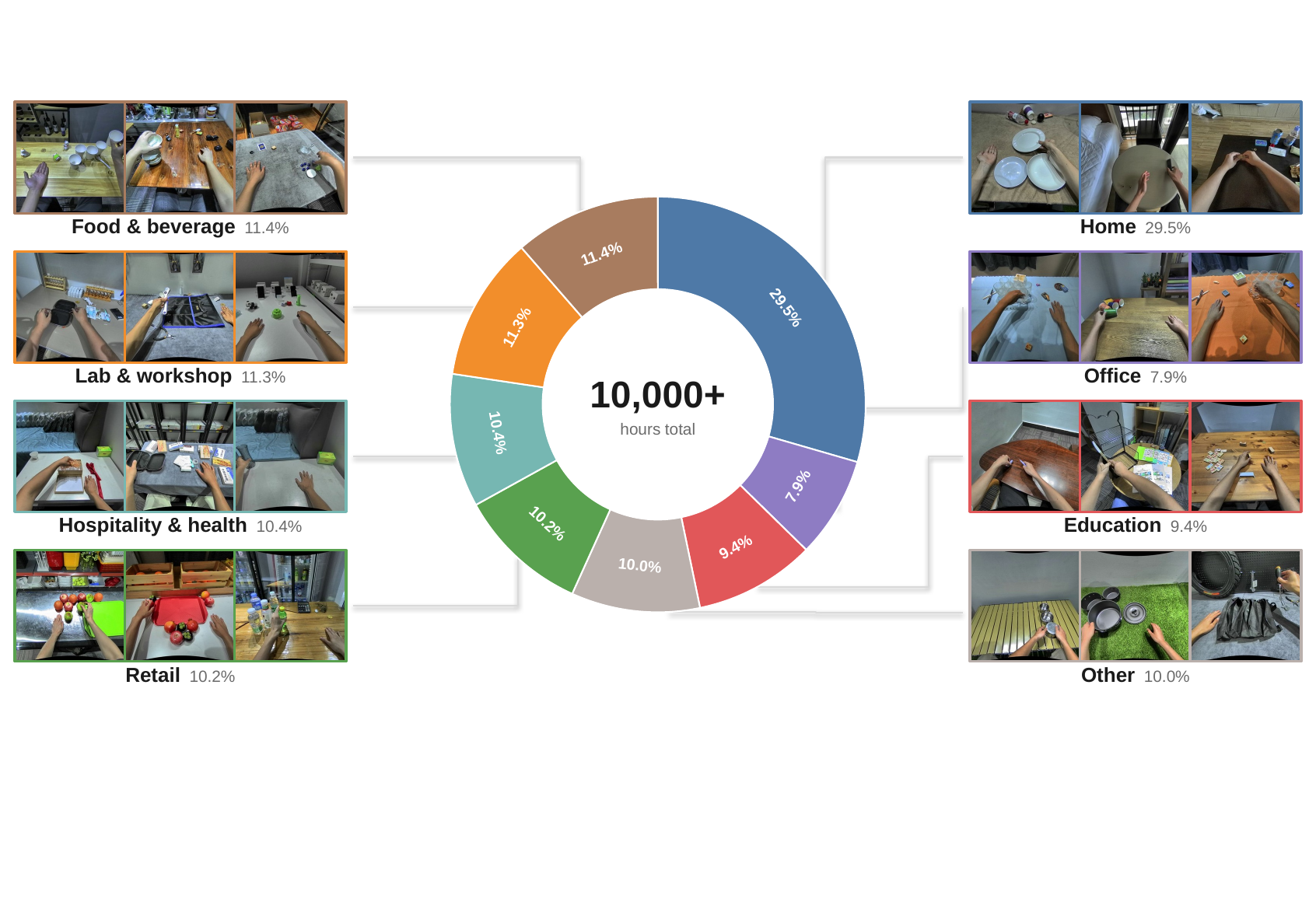}
  \caption{\textbf{Scene distribution of the IronMind pretraining corpus.} 
  The dataset comprises over 10,000 total hours of egocentric human video and heterogeneous robot manipulation data spanning diverse real-world environments, all expressed using a camera-space action representation.
  The donut chart shows the proportion of recording hours in each scene category, alongside representative frames. 
  Home environments account for the largest share; the remaining data cover food and beverage settings, laboratories and workshops, hospitality venues, retail spaces, educational facilities, offices, and other environments.}
  \label{fig:data-composition}
\end{figure}

\subsubsection{Camera-space action representation}
\label{sec:data-camspace}

Recent work highlights the scalability of egocentric human video and the effectiveness of the camera-space action representation~\citep{egoscale2026,egowam2026,li2025vitra}. 
In standard robot control, the torso frame is a natural reference for upper-limb manipulation. Egocentric video, however, records motion from the camera's perspective. Because low-cost recording hardware does not track the torso or spine, mapping human motion into a canonical robot torso frame requires assumptions about unobserved body kinematics (e.g., a rigid, vertical spine), which can introduce substantial retargeting noise~\citep{li2025vitra,qwenrobotmanip2026}.

We therefore bypass body retargeting by using a \emph{camera-space action representation}~\citep{qwenrobotmanip2026,joyai_ra05_2026,li2025vitra}. 
In addition to avoiding artifacts introduced by body retargeting, the camera-space action representation reduces sensitivity to \emph{camera mounting misalignments} on physical robots. On physical humanoid robots, small deviations in head-camera mounting can create mismatches between visual observations and actions expressed in a torso frame. Expressing observations and actions in the same camera space helps reduce these mismatches.

For human data, parametric 3D hand poses and trajectories are extracted in the recording camera space~\citep{mhr2025}. For robot data, measured trajectories are transformed into the onboard head-camera space using extrinsic calibration. These robot demonstrations come from heterogeneous non-target embodiments; no IRON-R01 demonstrations are included during pretraining.
To bridge structural differences between human hands and robot end-effectors, we use a unified action representation~\citep{qwenrobotmanip2026} that aligns action dimensions by their meaning across embodiments. Wrist and finger motions map to shared target-robot action dimensions.
This alignment is semantic rather than kinematic: after the camera-space transformations described above, corresponding motion quantities occupy consistent output dimensions even when embodiments differ in joint structure. The policy can therefore learn a single action interface across data sources.

\begin{figure}[t]
  \centering
  \includegraphics[width=1.0\linewidth]{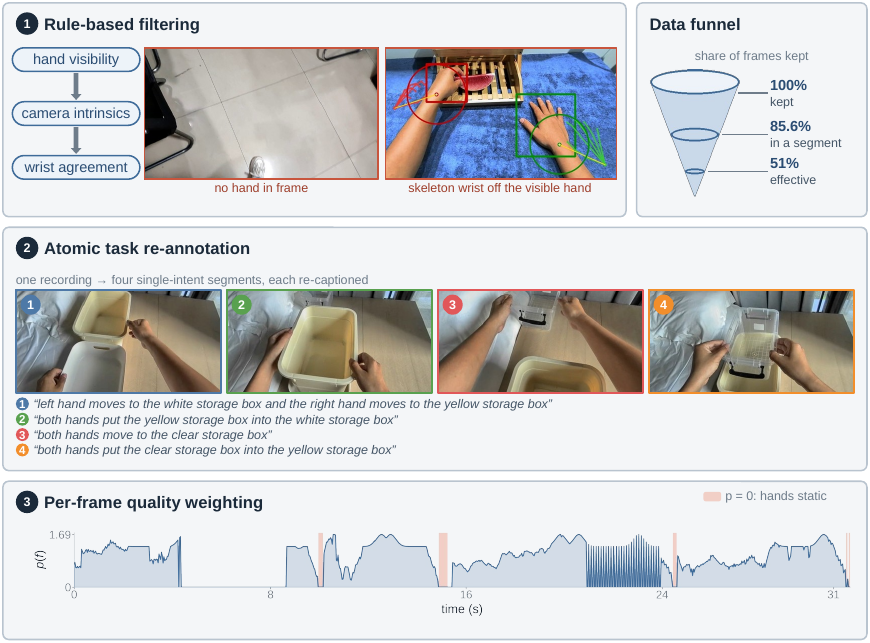}
  \caption{\textbf{Overview of the data curation pipeline with representative examples.} 
  Rule-based filtering rejects recordings that fail the quality criteria (two examples shown). 
  Atomic Re-annotation: Retained recordings are segmented into clips, each corresponding to a single task intent, and assigned refined captions. 
  Per-frame Quality Weighting: Frames receive quality scores according to Eq.~\ref{eq:frame-score} (plotted for a sample recording) to determine their sampling weights during training. 
  }
  \label{fig:curation-pipeline}
\end{figure}

\subsubsection{Data curation pipeline}
\label{sec:data-curation}

The shared action representation addresses embodiment mismatch, but it does not correct noisy trajectories or inaccurate language annotations.
Effective scaling of egocentric pretraining depends on the availability of correctly annotated, kinematically valid action trajectories, not simply on total video duration~\citep{egoscale2026}. 
Raw egocentric recordings often contain long periods of passive observation, mismatches between captions and actions (e.g., clips labeled ``pick up mug'' that instead show an object being released), and coarse task descriptions that do not distinguish individual sub-actions. 
To address these issues, we process all video streams through a unified multi-stage curation pipeline (Figure~\ref{fig:curation-pipeline}).

\textbf{Rule-based filtering.} We first apply rule-based checks to remove segments affected by perception and tracking failures. We reject frames with missing or invalid camera intrinsics or implausible 3D hand poses caused by hand-pose estimation errors or severe occlusions. 

\textbf{Atomic task annotation.} Coarse or missing captions provide insufficient supervision for policies that must follow fine-grained instructions. We segment continuous recordings into \emph{atomic task} units, each consisting of a short manipulation segment with a single intent, and caption each segment using a vision-language model (Qwen3.6-35B-A3B~\citep{qwenteam2026qwen36}) to describe the objects and actions involved. We mask action losses at segment boundaries to prevent training windows from mixing supervision across task intents.

\textbf{Kinematic velocity alignment.} Human hand movements in video are typically faster and more agile than the slower, smoother motions in robot teleoperation data. To reduce this difference, we compute the mean velocity for each source dataset split and apply temporal rescaling to better match human trajectory speeds to those of real-robot teleoperation data.

\textbf{Per-frame quality weighting.} In addition to binary filtering, we assign sampling weights to candidate frames within atomic segments using continuous quality measures. For each candidate frame $f$, a quality score $p(f)$ is computed over a short trailing window:
\begin{equation}
  p(f) \;=\; e(f)\,g(f) \cdot s_{\mathrm{cam}}(f)\,s_{\mathrm{move}}(f)\,b_{\mathrm{vel}}(f) \;\in\; [0,\, C_{\max}],
  \label{eq:frame-score}
\end{equation}
where $e(f)$ and $g(f)$ are binary gates for hand visibility and camera tracking stability, while $s_{\mathrm{cam}}(f)$, $s_{\mathrm{move}}(f)$, and $b_{\mathrm{vel}}(f)$ encode camera rotational stability, active hand translation, and a bonus for pauses at segment boundaries, respectively. %

The dataloader draws training samples using the cumulative distribution induced by $p(f)$ via inverse-transform sampling. This prioritizes frames with smooth motion and clear manipulation intent while retaining neighboring frames for context and avoiding costly dataset regeneration.

The stages serve distinct roles: filtering determines which observations are valid, atomic annotation defines the intended behavior, temporal rescaling reduces source-dependent speed differences, and quality weighting controls how often valid frames are sampled.

\FloatBarrier

%% file: sections/04_method/model.tex
\begin{figure}[!t]
  \centering
  \includegraphics[width=\linewidth]
{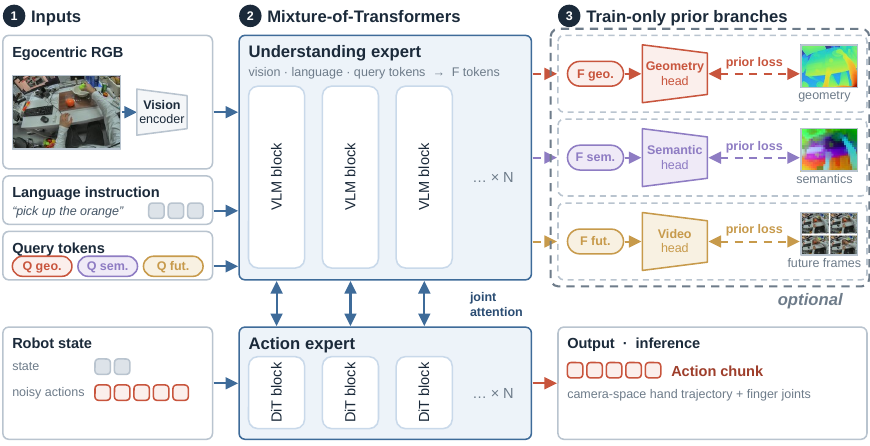} %
  \caption{\textbf{IronMind architecture.} A vision-language understanding
  expert takes an egocentric image, a language instruction, and optionally a set of
  query tokens as input. During training, the query tokens produce world prior features
  (geometry, semantics, future frames) and are supervised  using targets from pretrained models in each domain; the prior branches and prediction heads are removed at inference. The action expert attends
  to the understanding expert at shared attention layers and generates
  actions in the camera-space representation through flow-matching denoising, conditioned on
  robot state and action noise.}
  \label{fig:arch}
\end{figure}

\subsection{Model Architecture}
\label{sec:model}

IronMind is a vision-language-action (VLA) policy that maps egocentric visual observations and open-vocabulary language instructions to smooth, metric trajectories expressed using the camera-space action representation (Figure~\ref{fig:arch}). 
To achieve scalable pretraining while preserving reasoning and instruction-following capabilities, the architecture is designed to balance low-level trajectory regression with high-level vision-language understanding. 

IronMind addresses this via two core principles:
\textbf{(1) Dual-Expert Decoupling via Mixture-of-Transformers (MoT):} using separate experts for perception and reasoning and for flow-matching action decoding, with selective shared attention connecting the two; and 
\textbf{(2) Multimodal Prior Distillation via Latent Queries:} using auxiliary dynamics, geometry, and semantics objectives only during training.

\subsubsection{Dual-Expert Mixture-of-Transformers Architecture}
\label{sec:model-mot}

Standard single-stream VLA models append discrete or continuous action tokens directly to the vision-language context sequence. However, directly updating the vision-language trunk with gradients from low-level trajectory regression can interfere with its pretrained open-vocabulary representations. Following the Mixture-of-Transformers (MoT) paradigm~\citep{liang2024mot,internvla_a15_2026}, IronMind decouples perception and reasoning from action generation into two dedicated transformer streams with separate parameters:   

\textbf{Understanding Expert.} Initialized from a pretrained vision-language backbone, the understanding expert processes the head-camera RGB image and the natural-language instruction. It retains its native causal and cross-attention language decoding heads. This design helps preserve open-vocabulary grounding and instruction understanding during action pretraining.

\textbf{Action Expert.} Implemented as a Diffusion Transformer (DiT) variant trained with a flow-matching objective~\citep{lipman2023flow,black2024pi0}, the action expert models distributions over continuous action trajectories. During training, it conditions on state tokens and noisy action chunks to predict vector fields that transform noise into ground-truth action sequences. At inference, numerical integration of the predicted vector field over a small number of steps produces a temporally coherent action chunk. Predicting multi-step trajectories simultaneously encourages temporal smoothness and helps mitigate compounding execution errors during closed-loop rollout.

\textbf{Shared Attention Fusion.} IronMind connects the two experts through shared attention layers at selected depths~\citep{internvla_a15_2026}. At designated fusion layers throughout the transformer stack, understanding and action tokens participate in joint attention over the concatenated sequence, with stream-specific projection matrices applied to each stream. This selective cross-stream interaction allows the action expert to access fine-grained visual features and language context directly from the understanding backbone, while reducing direct interference with the pretrained vision-language representations.

\subsubsection{\texorpdfstring{Multimodal World Priors Supervision}{Multimodal World Priors Supervision}}
\label{sec:model-priors}

Robotic manipulation requires an understanding of depth, surface geometry, and changes in the scene during contact, all of which receive only indirect supervision from standard 3D trajectory regression. Following the latent-query formulation of InternVLA-A1.5 and EgoWAM~\citep{internvla_a15_2026,egowam2026}, IronMind optionally incorporates world priors during pretraining by adding a compact set of learnable \emph{world query} tokens to the understanding expert's prefix.

During training, the query tokens interact with the image and language context through self-attention in the understanding expert, producing compact latent representations of geometry, semantics, and future dynamics. These latent representations condition auxiliary heads that predict scene dynamics, geometry, and semantic features:

\emph{\textbf{(1) Spatial-Geometric Prior:}} the queries condition a prediction head supervised by dense depth representations~\citep{lin2025depthanything3}, encouraging the model to associate image locations with metric depth; 
\emph{\textbf{(2) Semantic Representation Prior:}} the queries condition a feature decoder that predicts semantic features from a self-supervised model~\citep{simeoni2025dinov3}, encouraging object representations that are less sensitive to appearance changes; and
\emph{\textbf{(3) Dynamic World-Model Prior:}} the latent queries condition a frozen video generation model~\citep{wan2025} to predict future variational autoencoder (VAE) latents, encouraging the understanding expert to encode contact dynamics and spatiotemporal changes. We also evaluate 3D motion flow as an alternative supervision target for this branch.

All external pretrained feature extractors, depth estimators, and video generation models remain frozen throughout policy pretraining. The auxiliary-loss gradients enter the understanding expert through the learnable query-token pathways. After pretraining, we remove the auxiliary decoder heads and target-generation networks and no longer compute the auxiliary losses. At inference, the action expert accesses the learned representations through shared attention. The learned query tokens are retained, while the auxiliary prediction branches are not executed during deployment.

%% file: sections/04_method/training.tex
\subsection{Training Paradigm}
\label{sec:training}

\begingroup\hbadness=2000
IronMind uses a two-stage training paradigm to transfer representations learned from heterogeneous pretraining data to the target robot. 
In the \textbf{large-scale pretraining stage}, we train the MoT policy primarily on large-scale egocentric human video, supplemented by heterogeneous robot teleoperation sequences from non-target embodiments. 
During this stage, we optimize the camera-space flow-matching action loss, optionally together with auxiliary objectives for depth, DINOv3 semantic features, and video dynamics. 
These world priors provide additional supervision during pretraining, encouraging the understanding expert to learn representations of scene geometry, semantics, and dynamics. 
Across the scaling experiments, the architecture, optimization recipe, and heterogeneous robot-data mixture remain fixed; additional experience is introduced through the human-video portion of the combined corpus. This design attributes differences across checkpoints primarily to data scale.
\par\endgroup

In the subsequent \textbf{post-training stage}, we adapt the pretrained policy to the target robot using a curated dataset of teleoperated trajectories. IRON-R01 demonstrations are introduced only at this stage.
The action interface retains the camera-space action representation: we transform robot trajectories into head-camera space using extrinsic calibration, allowing us to reuse the pretrained understanding expert and action head without changing the action representation (a torso-frame baseline is trained only for the comparison in Section~\ref{sec:exp-closedloop}). We fine-tune both experts on real-robot data to adapt the pretrained policy to IRON-R01.
Thus, post-training changes the embodiment while preserving pretrained action semantics.

%% file: sections/05_experiments/experiments.tex
\section{Experiments}
\label{sec:experiments}

Evaluating a Vision-Language-Action (VLA) manipulation policy directly on a real humanoid robot is time-intensive, safety-critical, and throughput-constrained. Real-robot supervision is naturally sparse, and the control loop between model decisions and physical outcomes is long. To systematically navigate this trade-off, we evaluate IronMind through a multi-tiered empirical framework that trades interaction fidelity against throughput. Our investigation centers on four primary dimensions: verifying policy capability scaling over total pretraining budgets up to 10,000 hours, examining optional auxiliary multimodal world-model priors, testing the camera-space action representation against a torso-frame baseline, and establishing real-robot success on IRON-R01 under out-of-distribution (OOD) scenarios.

\begin{figure}[t]
    \centering
    \includegraphics[
        page=1,
        trim=0mm 0mm 0mm 0mm,
        clip,
        width=\textwidth
    ]{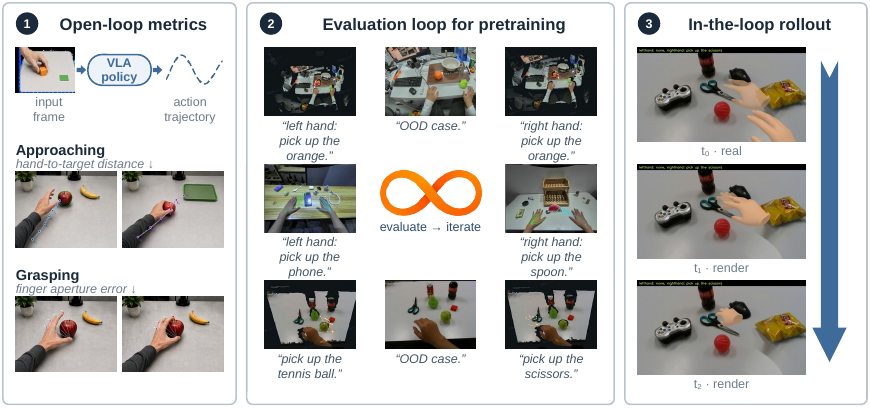}
    \caption{\textbf{Evaluation protocol.} \textbf{Left: Quantitative open-loop evaluation.} With the camera-space action representation, target approach is measured via $d_{p10}$ and grasp execution via finger aperture error $e_{\mathrm{FA}}$. \textbf{Middle: Evaluation loop for pretraining.} Held-out RealSense tabletop scenes with language prompts are used both to iterate pretraining and to score policy-in-the-loop rollouts. \textbf{Right: Policy-in-the-loop rollout.} Chunk $t_0$ is predicted from the observation; subsequent chunks $t_1,t_2$ re-render the predicted hand while the camera and physical scene stay fixed.}
    \label{fig:eval-protocol}
\end{figure}

\subsection{Implementation Details and Training Recipe}
\label{sec:exp-impl}
\label{sec:training-posttrain-curation}   %

\begingroup\hbadness=2000
All IronMind models are trained with AdamW~\citep{loshchilov2019decoupled} using a global batch size of $1024$ and a cosine learning-rate schedule that decays to $10\%$ of the peak rate. The reported pretraining budgets combine egocentric human video with heterogeneous robot data and contain no IRON-R01 demonstrations. Post-training uses approximately $40{,}000$ filtered teleoperated episodes collected on IRON-R01.
\par\endgroup

\subsection{Evaluation Protocols}
\label{sec:exp-methodology}

\begingroup\hbadness=10000
We adopt a hierarchical evaluation protocol with increasing interaction fidelity, spanning camera-space open-loop metrics, offline policy-in-the-loop rollouts, and real-robot execution (Figure~\ref{fig:eval-protocol}).
\par\endgroup

The three tiers isolate complementary sources of error. Open-loop evaluation measures the quality of a single predicted chunk without feedback, whereas policy-in-the-loop rollouts expose error accumulation when predictions are repeatedly conditioned on the model's own outputs. Real-robot evaluation then introduces factors absent from image-based rollouts, including controller error, camera motion, and physical contact. Using the same approaching metrics in the first two tiers makes their results directly comparable before moving to task-level success on IRON-R01.

\subsubsection{Open-Loop Evaluation}
\label{sec:exp-openloop}
Open-loop evaluation scores one predicted action chunk against the observed
scene without executing it. We collect all open-loop evaluation sets, and every
case starts from a real human hand reconstructed by the same front end that
labels the pretraining data, so the observation and the proprioceptive state
stay in one domain. We measure approaching and grasping quantitatively,
probe individual capabilities with case studies, and run a user study on
general actions.

\begin{figure}[t]
    \centering
    \includegraphics[width=\textwidth]{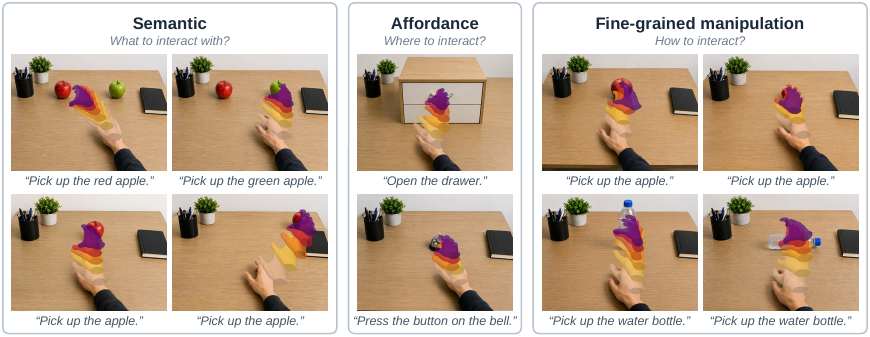}
    \caption{\textbf{Open-loop capability evaluation on OOD scenes.} Given an unseen RGB image and a text instruction, IronMind predicts open-loop hand trajectories (rendered hand meshes colored chronologically from light skin tone to purple). Visualizations illustrate fine-grained spatial grounding, including adjustments of grasp aperture and wrist orientation to target geometry.}
    \label{fig:capability-openloop}
\end{figure}

\paragraph{Quantitative Metrics.}
We score approaching and grasping on two self-collected sets.
\textbf{Approaching.} We instruct the model to reach for objects on a table-top. We capture $101$
RGB-D snapshots of a table-top with a RealSense D435, rearranging its $6$--$15$
everyday objects between shots, and annotate $23$ object categories with
captions and 3D point clouds segmented from the depth image. Each snapshot
also contains one or two real human hands resting $25$--$50$\,cm from the
objects. Pairing every hand with every unambiguous object in its snapshot
gives $970$ test cases ($492$ right-hand, $478$ left-hand) under $46$ prompts. Let $\mathcal{P}\subset\mathbb{R}^{3}$
be the target's point cloud and $\mathbf{x}_{t,k}$ be the predicted positions of
wrist and fingers. The per-step hand-to-target distance is
$d_t=\min_{k}\min_{\mathbf{p}\in\mathcal{P}}\|\mathbf{x}_{t,k}-\mathbf{p}\|_2$,
and we report three summaries of it over the chunk: $d_{p10}$, the $10$th
percentile of $\{d_t\}$ in cm, a robust closest-approach measure that is less sensitive to isolated steps; \emph{progress}, $(d_0-d_T)/d_0$, the share of the initial gap
closed by the last step; and
$\theta_{\mathrm{align}}=\angle(\bar{\mathbf{p}}-\mathbf{x}_{0,\mathrm{wrist}},\,
\mathbf{x}_{T,\mathrm{wrist}}-\mathbf{x}_{0,\mathrm{wrist}})$, the angle
between the direction to the target centroid $\bar{\mathbf{p}}$ and the
wrist's net displacement, $0^\circ$ meaning the hand heads straight for the
target. Lower is better for $d_{p10}$ and $\theta_{\mathrm{align}}$, higher
for progress.
\textbf{Grasping.} The grasping set contains $120$ grasp events from $10$ self-collected scenes, one target object each,
with human-reviewed closure-onset anchors. Mean finger aperture is
$a_t=\frac{1}{4}\sum_{j=1}^{4}\|\mathbf{x}_{t,\mathrm{thumb}}-\mathbf{x}_{t,j}\|_2$
over the four fingertips. The ground-truth segment from the anchor, where the
hand starts to close, to full closure is resampled to the length of one action
chunk, the policy predicts one chunk from the anchor observation, and
$e_{\mathrm{FA}}$ is the mean absolute difference between the predicted and
ground-truth aperture profiles over that chunk (mm), measuring how closely the predicted closure follows the demonstration.

\paragraph{Case Studies across Capabilities.}
Beyond the aggregate metrics, we probe individual capabilities with case
studies on self-collected out-of-distribution scenes: semantic grounding
(\emph{what} to interact with), affordance (\emph{where} to interact) and
fine-grained manipulation (\emph{how} to interact). For each case we render
the predicted hand trajectory in 2D and 3D and inspect where each capability succeeds or fails (Figure~\ref{fig:capability-openloop}).

\paragraph{User Study on General Actions.}
For general hand actions beyond reaching and grasping, quantitative metrics do
not adequately capture whether a predicted movement is plausible, so we run a
user study. For each held-out scene we prompt the models under comparison
with the annotated instruction and render the predicted hand actions onto the
frames. A trial shows the ground-truth demonstration next to the unlabeled,
shuffled renderings, and the participant picks the better prediction or a tie.
The study compares the 10,000\,h and 250\,h checkpoints on $20$ held-out trials, each judged by $5$ participants, yielding $100$ votes. We compute win rate over decided votes and report ties in Table~\ref{tab:human-pref}.

\subsubsection{Policy-in-the-Loop Rollout}
To probe sequential error accumulation over multi-chunk execution without running the policy on a robot, we run policy-in-the-loop rollouts on the same $970$ approaching cases. We chain three relative chunks. After chunk $k$, the predicted hand is rendered back into the observation for chunk $k+1$; camera pose and scene geometry stay fixed (Figure~\ref{fig:eval-protocol}, right). We report the approaching metrics of the open-loop protocol, $d_{p10}$, progress and $\theta_{\mathrm{align}}$, over all $970$ cases (Table~\ref{tab:semi-closed}). Because three chunks are long enough to close a substantial fraction of the $25$--$50$\,cm start gap, $d_{p10}$ here is a reach score rather than a restatement of $d_0$.

\subsubsection{Closed-Loop Real-Robot Evaluation}
Real-robot evaluations use an IRON-R01 unit distinct from the one used for post-training data collection, with the policy running in closed loop at $10$\,Hz under contact dynamics and sensor feedback.

\subsection{Pretraining Experience Scaling}
\label{sec:exp-scaling}

\begin{figure}[t]
  \centering
  \includegraphics[width=0.85\linewidth]{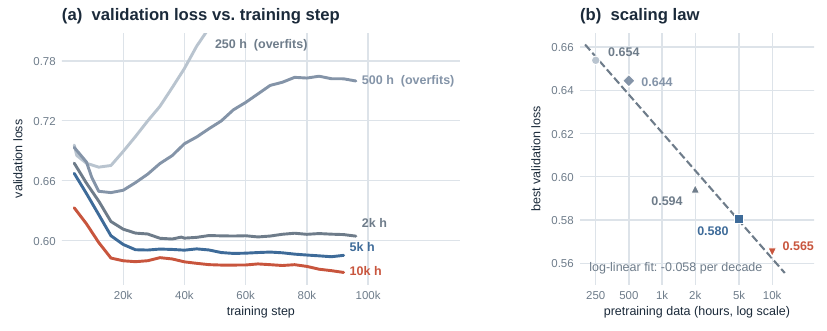}
  \caption{\textbf{Pretraining data scaling.} Pretraining validation loss exhibits a smooth log-linear scaling trend across total pretraining budgets (250\,h to 10,000\,h) with a slope of $-0.058$ per decade of hours, with larger budgets also improving open-loop target approach and grasp closure.}
  \label{fig:scaling}
\end{figure}

Across pretraining budgets from 250 to 10,000 hours, IronMind's validation loss follows an approximately log-linear scaling law (Figure~\ref{fig:scaling}). The scaling fits a log-linear slope of $-0.058$ per decade of pretraining hours without exhibiting loss saturation up to 10,000 hours.

\begin{table}[t]
\centering
\footnotesize
\setlength{\tabcolsep}{3.5pt}
\begin{minipage}[t]{0.55\textwidth}
\centering
\captionof{table}{\textbf{Open-loop performance vs. pretraining budget\\} (one chunk).}
\label{tab:open-loop-scaling}
\begin{tabular}{lcccc}
\toprule
Budget & $d_{p10}$\,(cm)\,$\downarrow$ & Prog.\,(\%)\,$\uparrow$ & $\theta_{\mathrm{align}}$\,($^\circ$)\,$\downarrow$ & $e_{\mathrm{FA}}$\,(mm)\,$\downarrow$ \\
\midrule
250\,h     & $37.8$          & $14.7$           & $47.8$          & $19.97$ \\
500\,h     & $33.4$          & $28.1$           & $41.7$          & $19.54$ \\
2,000\,h   & $32.0$          & $30.7$           & $36.0$          & $23.83$ \\
5,000\,h   & $30.8$          & $34.0$           & $41.2$          & $20.33$ \\
10,000\,h  & $\mathbf{25.8}$ & $\mathbf{44.9}$  & $\mathbf{26.0}$ & $\mathbf{15.66}$ \\
\bottomrule
\end{tabular}
\end{minipage}\hfill
\begin{minipage}[t]{0.435\textwidth}
\centering
\captionof{table}{\textbf{Policy-in-the-loop performance vs. pretraining budget} (three chunks, $970$ cases).}
\label{tab:semi-closed}
\begin{tabular}{lccc}
\toprule
Budget & $d_{p10}$\,(cm)\,$\downarrow$ & Prog.\,(\%)\,$\uparrow$ & $\theta_{\mathrm{align}}$\,($^\circ$)\,$\downarrow$ \\
\midrule
250\,h     & $32.0$          & $21.4$          & $46.1$ \\
500\,h     & $25.6$          & $32.2$          & $39.5$ \\
2,000\,h   & $23.0$          & $39.0$          & $33.9$ \\
5,000\,h   & $22.7$          & $39.1$          & $36.3$ \\
10,000\,h  & $\mathbf{14.6}$ & $\mathbf{49.3}$ & $\mathbf{21.6}$ \\
\bottomrule
\end{tabular}
\end{minipage}
\end{table}

Larger pretraining budgets also improve open-loop spatial and grasping
behavior (Table~\ref{tab:open-loop-scaling}). Scaling pretraining data from
250\,h to 10,000\,h reduces $d_{p10}$ from $37.8$ to $25.8$\,cm
($31.7\%$), increases approach progress from $14.7\%$ to $44.9\%$, and
lowers the finger-aperture error $e_{\mathrm{FA}}$ from $19.97$ to
$15.66$\,mm. The intermediate budgets are not strictly ordered by every
metric, but target approach improves monotonically with data, and the largest
budget is best in all four columns.
The 10,000-hour checkpoint also performs best in policy-in-the-loop rollouts
(Table~\ref{tab:semi-closed}): from 250\,h to 10,000\,h, progress
rises from $21.4\%$ to $49.3\%$, alignment falls from $46.1^\circ$ to
$21.6^\circ$, and $d_{p10}$ from $32.0$ to $14.6$\,cm, while 5,000\,h
and 2,000\,h stay nearly tied ($39.1\%$ vs.\ $39.0\%$ progress).
In the blind user study on
held-out egocentric frames (Table~\ref{tab:human-pref}), raters prefer the
10,000\,h pretrained checkpoint over the 250\,h baseline in $47$ of the
$70$ decided votes, a \textbf{$67.1\%$ win rate}; the remaining $30$ of the
$100$ votes are ties.

\begin{table}[ht]
\centering
\caption{\textbf{Human preference win rate vs. pretraining data budget.}
The blind user study in Section~\ref{sec:exp-openloop} covers $20$ held-out trials:
$100$ votes: $30$ ties and $70$ decisions; win rate is computed over decided votes.}
\label{tab:human-pref}
\small
\setlength{\tabcolsep}{12pt}
\begin{tabular}{lcc}
\toprule
Pretraining Budget & Votes & Win Rate $\uparrow$ \\
\midrule
250\,h     & $23$ & $32.9\%$ \\
10,000\,h  & $\mathbf{47}$ & $\mathbf{67.1\%}$ \\
\bottomrule
\end{tabular}
\end{table}

\subsection{\texorpdfstring{World Priors Ablation}{World Priors Ablation}}
\label{sec:exp-ablations}

\begin{figure}[!ht]
  \centering
  \includegraphics[width=0.9\textwidth]{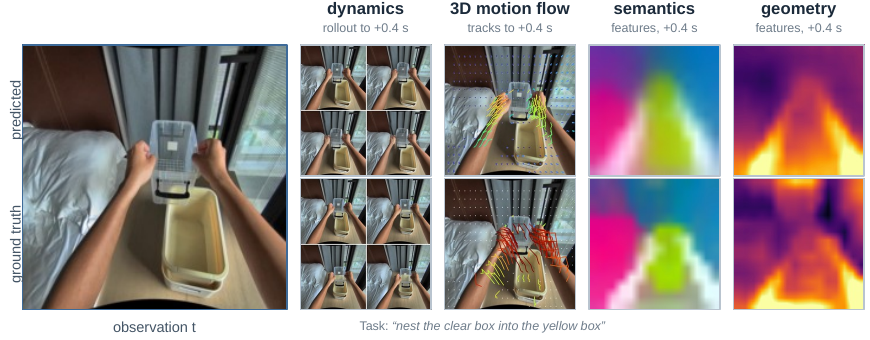}
  \caption{\textbf{What the world prior branches predict from one observation.}
  The left panel shows observation $t$; for each branch the prediction
  (top) sits above the ground truth (bottom) in the same rendering. The panels show future-frame dynamics, 3D motion flow, semantic features, and geometry targets over a $0.4$\,s horizon.
  }
  \label{fig:wm-prior-targets}
\end{figure}

\begin{table}[!ht]
\centering
\caption{\textbf{World-prior ablation on the open-loop approaching set.} Results are averaged over several checkpoints from each pretraining run. Lower is better for $d_{p10}$ and $\theta_{\mathrm{align}}$; higher is better for progress. }
\label{tab:wm-prior-ablation}
\small
\begin{tabular}{lccc}
\toprule
Prior Configuration & $d_{p10}$ (cm) $\downarrow$ & Prog.\,(\%) $\uparrow$ & $\theta_{\mathrm{align}}$ (°) $\downarrow$ \\
\midrule
\multicolumn{4}{l}{\textit{Reference configurations}} \\
no world-prior supervision (baseline) & $27.6$ & $40.9$ & $34.8$ \\
three priors (default: future target, $0.4$\,s) & $\mathbf{22.6}$ & $\mathbf{53.4}$ & $\mathbf{26.5}$ \\
\midrule
\multicolumn{4}{l}{\textit{Prior composition}} \\
\quad without the semantics prior     & $24.1$ & $50.2$ & $29.2$ \\
\quad without the depth prior         & $25.7$ & $46.1$ & $31.1$ \\
\quad without the dynamics prior      & $24.1$ & $51.1$ & $26.8$ \\
\midrule
\multicolumn{4}{l}{\textit{Dynamics supervision}} \\
\quad target: current frame          & $27.3$ & $42.7$ & $33.3$ \\
\quad horizon $0.2$\,s                & $25.8$ & $45.7$ & $30.5$ \\
\quad horizon $0.8$\,s                & $24.8$ & $49.3$ & $28.3$ \\
\quad dynamics target: 3D motion flow & $25.2$ & $47.1$ & $28.6$ \\
\bottomrule
\end{tabular}
\end{table}

All arms use the same pretraining recipe and corpus and differ from the three-prior configuration by a single configuration change. The three priors are the semantic, depth, and dynamics branches described in Section~\ref{sec:model-priors}, supervised by DINOv3, Depth-Anything-3-Base, and the frozen Wan2.2 video generator, respectively. Table~\ref{tab:wm-prior-ablation} reports results on the $970$-case approaching set described in Section~\ref{sec:exp-openloop}. Each entry is averaged over several checkpoints from the same training run. Because the differences are only a few centimeters, we interpret them as tendencies rather than conclusive effects. One arm replaces the frozen video generator with an EgoWAM-style 3D motion-flow head~\citep{egowam2026}. As illustrated in Figure~\ref{fig:wm-prior-targets}, this flow-matching decoder predicts the camera-space displacement of query points over the prediction horizon, using offline depth and tracking labels for supervision.

\textbf{The full three-prior configuration performs best across all three reported metrics.} Relative to the no-prior baseline, the full configuration reduces $d_{p10}$ from $27.6$ to $22.6$\,cm, increases
progress from $40.9\%$ to $53.4\%$ and improves alignment from $34.8^\circ$ to
$26.5^\circ$ (Table~\ref{tab:wm-prior-ablation}). Removing any one branch increases $d_{p10}$ by $1.5$--$3.1$\,cm, with the largest change observed when depth is removed, which is consistent with contributions from all three branches. Replacing future prediction with current-frame reconstruction largely reverses the gain ($27.3$\,cm, $42.7\%$), consistent with future targets providing useful information beyond current-frame feature supervision. The default horizon of approximately $0.4$\,s also performs better than $0.2$ or $0.8$\,s, possibly reflecting a trade-off between predictive information and uncertainty at longer horizons. At this training budget, the frozen video generator shows a modest advantage over the 3D motion-flow head as a dynamics target; the sparser flow labels may partly explain this difference.

\subsection{Real-Robot Experiments}
\label{sec:exp-closedloop}

\begin{figure}[!ht]
    \centering
     \includegraphics[width=\linewidth]{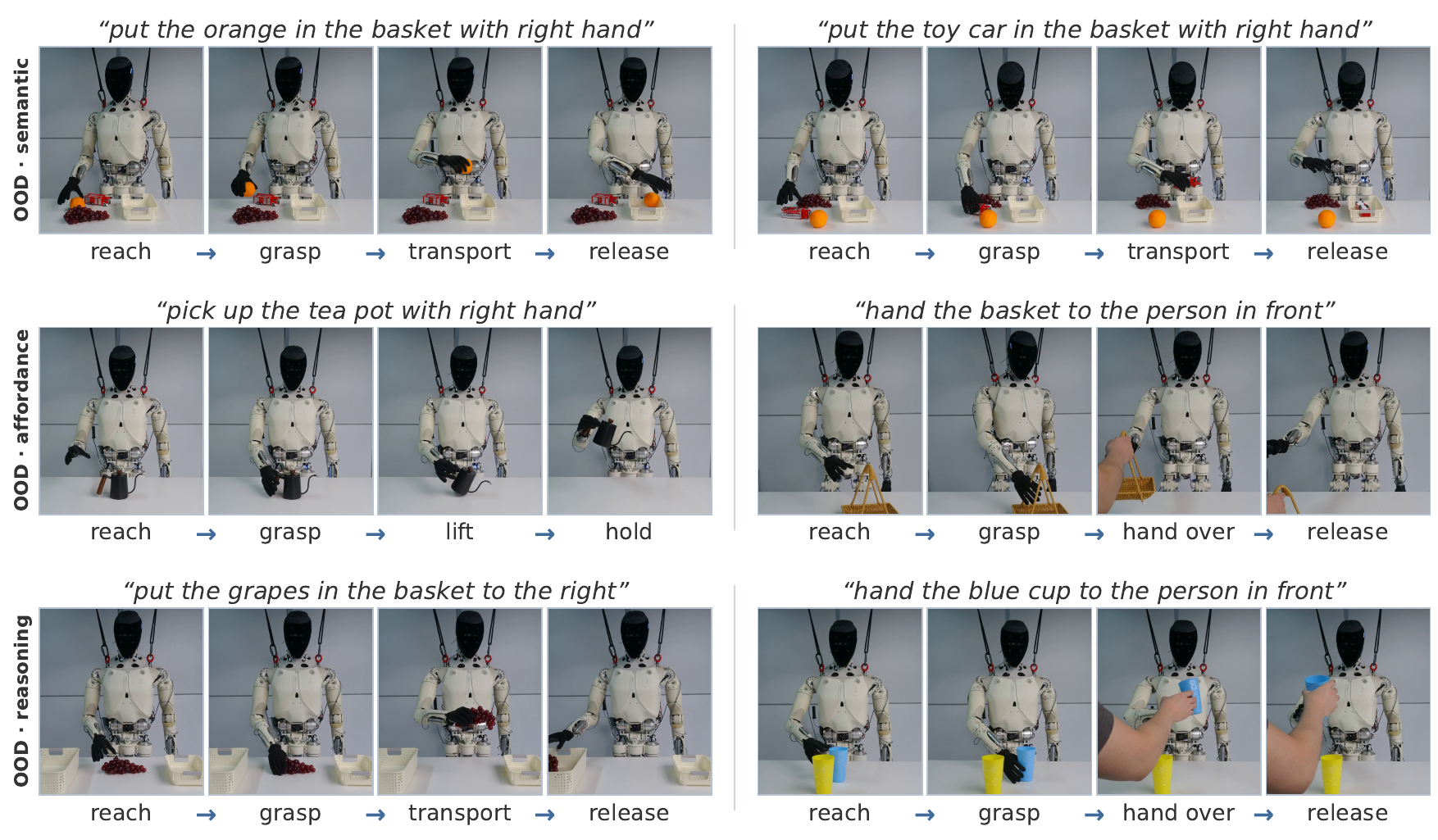}
    \caption{\textbf{Real-robot out-of-distribution (OOD) evaluation on IRON-R01.} Success and progress rates are evaluated across three out-of-distribution categories (Semantic, Affordance, and Reasoning OOD).}
    \label{fig:closedloop-real}
\end{figure}

\begin{figure}[!ht]
    \centering
    \includegraphics[width=1.0\linewidth]
        {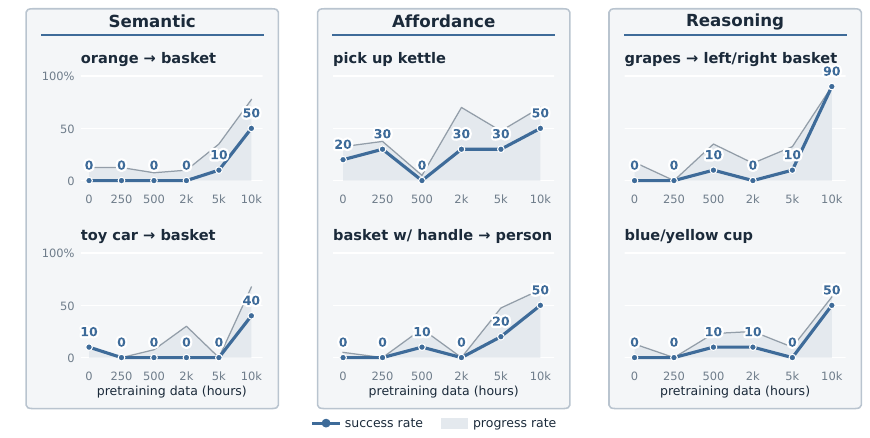}
    \caption{\textbf{Scaling effect on real-robot OOD success and progress rates.} Success (blue) and milestone progress (gray) are reported for six OOD tasks grouped into Semantic, Affordance, and Reasoning categories. At smaller budgets (0--2,000\,h), performance is generally low and non-monotonic across tasks. The clearest improvement occurs between 5,000 and 10,000\,h, when success increases on all six tasks---for example, from $10\%$ to $90\%$ for grapes to the left/right basket and from $0\%$ to $50\%$ for the blue/yellow cup---showing an overall positive scaling trend despite fluctuations at smaller scales.}
    \label{fig:real-robot-scaling}
\end{figure}

We evaluate IronMind in closed loop on IRON-R01 at $10$\,Hz. To test zero-shot generalization and spatial robustness, we use out-of-distribution (OOD) settings with object instances, affordances, and reasoning prompts not seen during post-training (Figure \ref{fig:closedloop-real}). For each setting, we report binary task success and a milestone-based progress score ($25\%$, $50\%$, $75\%$, or $100\%$ completion) over 10 trials.
Success requires completion of the full instructed task, whereas progress credits completed stages such as reaching, grasping, transporting, or releasing. Reporting both distinguishes immediate failures from executions that complete most of the sequence but miss the final condition.

We categorize real-robot out-of-distribution (OOD) evaluations into three groups: \textbf{semantic generalization}, \textbf{affordance adaptation}, and \textbf{spatial reasoning}. 
Semantic generalization tests grounding of unseen object instances and appearances, such as placing an orange or a toy car into a basket under visual clutter. 
Affordance adaptation tests interactions that require unseen grasp strategies, such as grasping a kettle by its handle rather than its body, or lifting a basket by its handle to hand it to a person. %
Spatial reasoning tests relational and attribute-based instructions (e.g., \emph{``put the grapes in the left basket''} or \emph{``pick up the blue cup rather than the yellow cup''}).

Real-robot results on IRON-R01 are consistent with the pretraining scaling trends observed in open-loop evaluation. Across OOD task categories (Figure~\ref{fig:real-robot-scaling}), the average success rate stays below $12\%$ for every training budget up to 5,000 hours and rises above $40\%$ ($40$--$90\%$ per task) at 10,000 hours. The increase is strongest on the Reasoning and Semantic OOD tasks and weaker on the Affordance OOD tasks. To isolate the effect of data scale, these experiments do not use world-prior supervision. 

We also compare IronMind variants with the external VLA baseline InternVLA-A1.5~\citep{internvla_a15_2026}. We take the public weights from InternVLA-A1.5 and post-train it on the same post-training data as IronMind. We then compare it with the 10,000-hour IronMind model on the same OOD tasks as in Figure~\ref{fig:closedloop-real}. The results are shown in Table~\ref{tab:a15-baseline}. IronMind (with the camera-space action representation and without world-prior supervision) obtains the highest overall success and progress rates under this evaluation setup. However, IronMind uses a shared unified action space for pretraining and post-training, while the action space is re-initialized for InternVLA-A1.5. %

\begin{table}[!t]
\centering
\small
\setlength{\tabcolsep}{7pt} %
\caption{\textbf{Real-robot ablations.} Results are averaged over the six OOD tasks of Figure~\ref{fig:closedloop-real} (10 trials each), using the same post-training data and compute budget. Parentheses report percentage-point drops relative to the first row. }
\label{tab:a15-baseline}
\begin{tabular}{lcc}
\toprule
Model variant & Success $\uparrow$ & Progress $\uparrow$ \\
\midrule
IronMind (ours, camera-space, no world-prior supervision) & \textbf{55.0\%} & \textbf{71.3\%} \\
IronMind (ours, torso-frame baseline) & 26.7\%\,{\scriptsize($\downarrow$28.3)} & 46.2\%\,{\scriptsize($\downarrow$25.1)} \\
IronMind (ours, with world-prior supervision) & 31.7\%\,{\scriptsize($\downarrow$23.3)} & 53.3\%\,{\scriptsize($\downarrow$18.0)} \\
\midrule
InternVLA-A1.5 ~\citep{internvla_a15_2026} & 30.0\%\,{\scriptsize($\downarrow$25.0)} & 47.8\%\,{\scriptsize($\downarrow$23.5)} \\
\bottomrule
\end{tabular}
\end{table}

Ablation experiments comparing different variants of IronMind (Table \ref{tab:a15-baseline}) show two patterns regarding representation space and supervision objectives.
First, the camera-space action representation performs better than the torso-frame baseline. To ensure a rigorous and fair comparison, both models were trained on the same 10,000-hour dataset (albeit transformed to their respective action representations). Each model used its assigned action representation consistently during both pretraining and post-training. These results suggest that a camera-space action representation can improve generalization in egocentric pretraining for humanoid robot manipulation.
Although world-prior supervision improves some open-loop metrics, its real-robot variant only modestly outperforms InternVLA-A1.5 and remains below IronMind without world priors. This may reflect a mismatch between auxiliary prediction objectives and contact-rich closed-loop control.

%% file: sections/06_conclusion/limitations.tex
\section{Limitations}
\label{sec:limitations}

Several limitations remain.
First, the corpus remains concentrated in homes, kitchens, and workshops, limiting extreme scene diversity.
Second, the camera-space action representation reduces but does not eliminate target-robot dependence, still requiring camera extrinsic calibration and post-training on real-robot data.
The scaling study covers one architecture and data mixture, while real-robot evaluation covers one embodiment, six OOD tasks, and 10 trials per task; longer-horizon and broader contact-rich tasks remain untested. Future work will further examine how gains from world-prior supervision transfer from open-loop evaluation to closed-loop deployment.
Finally, comparisons beyond InternVLA-A1.5, including DreamZero~\citep{dreamzero2026}, remain future work.

%% file: sections/06_conclusion/conclusion.tex
\section{Conclusion}
\label{sec:conclusion}

We presented IronMind, a VLA system pretrained on large-scale egocentric human and heterogeneous robot data for humanoid dexterous manipulation. Its unified camera-space action representation bridges human observation and robot execution without explicit embodiment retargeting. A curated data engine supports a dual-expert Mixture-of-Transformers policy with flow-matching action generation and optional multimodal world priors. Validation loss scales approximately log-linearly from 250 to 10,000 pretraining hours. Evaluation spans open-loop prediction, policy-in-the-loop rollouts, and closed-loop robot execution. After post-training, the 10,000-hour model achieves $55.0\%$ success across six out-of-distribution tasks and outperforms the torso-frame baseline ($55.0\%$ vs. $26.7\%$). Together, these results support camera-space pretraining as a scalable foundation for humanoid manipulation. Future work will scale data and embodiments and explore longer-horizon tasks, reinforcement learning, and human intervention.